%% file: main.tex
\documentclass[letterpaper, 10 pt, conference]{ieeeconf}  % Comment this line out if you need a4paper

\IEEEoverridecommandlockouts                              % This command is only needed if 
\usepackage{graphics} % for pdf, bitmapped graphics files
\usepackage{epsfig} % for postscript graphics files
\usepackage{amsmath} % assumes amsmath package installed
\usepackage{amssymb}  % assumes amsmath package installed
\usepackage{algorithm}
\usepackage{algorithmic}

\usepackage{environ}
\usepackage{xcolor}
\usepackage[colorlinks]{hyperref}

\usepackage{array}
\newcolumntype{P}[1]{>{\centering\arraybackslash}p{#1}}

\usepackage{array}
\usepackage{multirow}
\usepackage[nocompress]{cite}

\usepackage[utf8]{inputenc}

\DeclareUnicodeCharacter{2212}{-}

\newcommand\MyBox[2]{
  \fbox{\lower0.75cm
    \vbox to 1.5cm{\vfil
      \hbox to 1.5cm{\hfil\parbox{1.2cm}{#1\\#2}\hfil}
      \vfil}%
  }%
}

\title{\LARGE \bf
Forecasting Time-to-Collision from Monocular Video: \\ Feasibility, Dataset, and Challenges}

\author{Aashi Manglik, Xinshuo Weng, Eshed Ohn-Bar and Kris M. Kitani$^{1}$% <-this % stops a space
\thanks{$^{1}$Authors are affiliated with the Robotics Institute, Carnegie Mellon University, Pittsburgh, PA 15213, USA.  {\tt\small \{amanglik, xinshuow, eohnbar, kkitani\}@andrew.cmu.edu}. Eshed Ohn-Bar is now at the Max Planck Institute for Intelligent Systems.%
}}

\begin{document}

\maketitle
\thispagestyle{empty}
\pagestyle{empty}

%%%%%%%%%%%%%%%%%%%%%%%%%%%%%%%%%%%%%%%%%%%%%%%%%%%%%%%%%%%%%%%%%%%%%%%%%%%%%%%%
\begin{abstract}
We explore the possibility of using a single monocular camera to forecast the time to collision between a suitcase-shaped robot being pushed by its user and other nearby pedestrians. We develop a purely image-based deep learning approach that directly estimates the time to collision without the need of relying on explicit geometric depth estimates or velocity information to predict future collisions. While previous work has focused on detecting immediate collision in the context of navigating Unmanned Aerial Vehicles, the detection was limited to a binary variable (\emph{i.e.}, collision or no collision). We propose a more fine-grained approach to collision forecasting by predicting the exact time to collision in terms of milliseconds, which is more helpful for collision avoidance in the context of dynamic path planning. To evaluate our method, we have collected a novel dataset of over 13,000 indoor video segments each showing a trajectory of at least one person ending in a close proximity (a near collision) with the camera mounted on a mobile suitcase-shaped platform. Using this dataset, we do extensive experimentation on different temporal windows as input using an exhaustive list of state-of-the-art convolutional neural networks (CNNs). Our results show that our proposed multi-stream CNN is the best model for predicting time to near-collision. The average prediction error of our time to near-collision is $0.75$ seconds across the test videos. The project webpage can be found at \url{https://aashi7.github.io/NearCollision.html}.
\end{abstract}

% performance in terms of latency and accuracy 
% \begin{comments}
% You may also put your comments in red using \\begin\{comments\}
% \\end\{comments\}
% \end{comments}

%%%%%%%%%%%%%%%%%%%%%%%%%%%%%%%%%%%%%%%%%%%%%%%%%%%%%%%%%%%%%%%%%%%%%%%%%%%%%%%%
\section{INTRODUCTION}\label{intro}
\input{sections/introduction.tex}

\section{Related Work}\label{related_work}
\input{sections/related_work.tex}

\section{DATASET}\label{dataset}
\input{sections/dataset.tex}

\section{APPROACH}\label{approach}
\input{sections/approach.tex}

\section{EXPERIMENTAL EVALUATION}\label{experiments}
\input{sections/experimental_evaluation.tex}

% \section{RESULTS}
% \input{sections/results.tex}

\section{Discussion}
%% This question is now missing from the introduction 
We return to the question posed in introduction, 'Is it possible to predict the time to collision from a single camera?'. The answer is that the proposed model is able to leverage spatio-temporal cues for predicting the time to collision. Also, we observed that the multi-stream network of shared weights performed the task of collision forecasting better than I3D on the proposed dataset. 
%One of the major boosts to multi-stream VGG is the reuse of VGG-16 network pretrained on PASCAL VOC classification dataset which contains cimages of person class whereas I3D is pretrained Kinetics Human Action Video. 

Regarding the temporal window of input history, it is evident that using a sequence of images has a considerable benefit over prediction from single frame only. Though a temporal window incorporating 0.5 seconds of history performed best for our task, we do not observe a piecewise monotonic relation between input frames and error in prediction. This observation aligns with the performance of constant velocity model where accuracy in prediction does not necessarily increase or decrease with the temporal footprint of past trajectory. 

During data collection, the pushing speed of suitcase varies between $0.2-1.5$ meters per second which is similar in range to human walking speed. We understand that the proposed model might not generalize well when there are significant changes in camera's height, pushing speed of suitcase or structure of the scenes in comparison to provided dataset. In the scenarios where assistive robot operates in a constrained domain like museums and airports, the proposed approach will be suitable for predicting time to collision requiring only a low-cost monocular camera. We believe that the proposed approach could be extended for outdoor use given enough training data of outdoor scenes. 

One of the scenarios where the current approach fails is when a person is moving away from the user in the same direction. The pedestrian detector detects the person and thus the image is passed through the proposed multi-stream network. The network outputs an inaccurate estimate of time-to-collision within the trained forecast horizon. In the future work, we can instead try to output two values - mean and variance in time-to-collision. The variance can act as an indicator on how confident or reliable is the prediction. To train the output tuple of mean and variance, the negative of gaussian log-likelihood loss can be minimized over the training data which includes such failure cases.
%% assisitive robots - museum, airports wotk in restricted domains, so deep learning would perform better given enough data 

% A conclusion section is not required. Although a conclusion may review the main points of the paper, do not replicate the abstract as the conclusion. A conclusion might elaborate on the importance of the work or suggest applications and extensions. 

\addtolength{\textheight}{-12cm}   % This command serves to balance the column lengths

%%%%%%%%%%%%%%%%%%%%%%%%%%%%%%%%%%%%%%%%%%%%%%%%%%%%%%%%%%%%%%%%%%%%%%%%%%%%%%%%

%%%%%%%%%%%%%%%%%%%%%%%%%%%%%%%%%%%%%%%%%%%%%%%%%%%%%%%%%%%%%%%%%%%%%%%%%%%%%%%%



\section*{ACKNOWLEDGMENT}
This work was sponsored in part by NIDILRR (90DPGE0003), JST CREST (JPMJCR14E1) and NSF NRI (1637927).

%%%%%%%%%%%%%%%%%%%%%%%%%%%%%%%%%%%%%%%%%%%%%%%%%%%%%%%%%%%%%%%%%%%%%%%%%%%%%%%%

\bibliographystyle{IEEEtran}
\bibliography{IEEEabrv,references}

\end{document}

%% file: sections/introduction.tex
%%%%%%%%%%% First paragraph - Motivating time to collision %%%%%%%%%%%%%
Automated collision avoidance technology is an indispensable part of mobile robots. As an alternative to traditional approaches using multi-modal sensors, purely image-based collision avoidance strategies \cite{gandhi, DroNet, chan2016, Fu2019} have recently gained attention in robotics. These image-based approaches use the power of large data to detect immediate collision as a binary variable - collision or no collision. In this work, we propose a more fine-grained approach to predict the exact time to collision from images, with a much longer prediction horizon.  

% Automated collision avoidance technology is an indispensable part of autonomous assistive mobile robots especially in the presence of people. For example, \cite{BBeep} proposes a collision avoidance system which could be used to give simple feedback to the user by emitting a sound alarm expecting the humans to avoid a future collision. Also, the output of the collision avoidance system could be used directly for reactive control \cite{gandhi}, \cite{DroNet}. In this work, we explore the use of a convolutional neural network-based approach that directly regresses the time to near-collision from images. 

   \begin{figure}[t]
      \centering
      \includegraphics[height=3.8cm, width=\columnwidth]{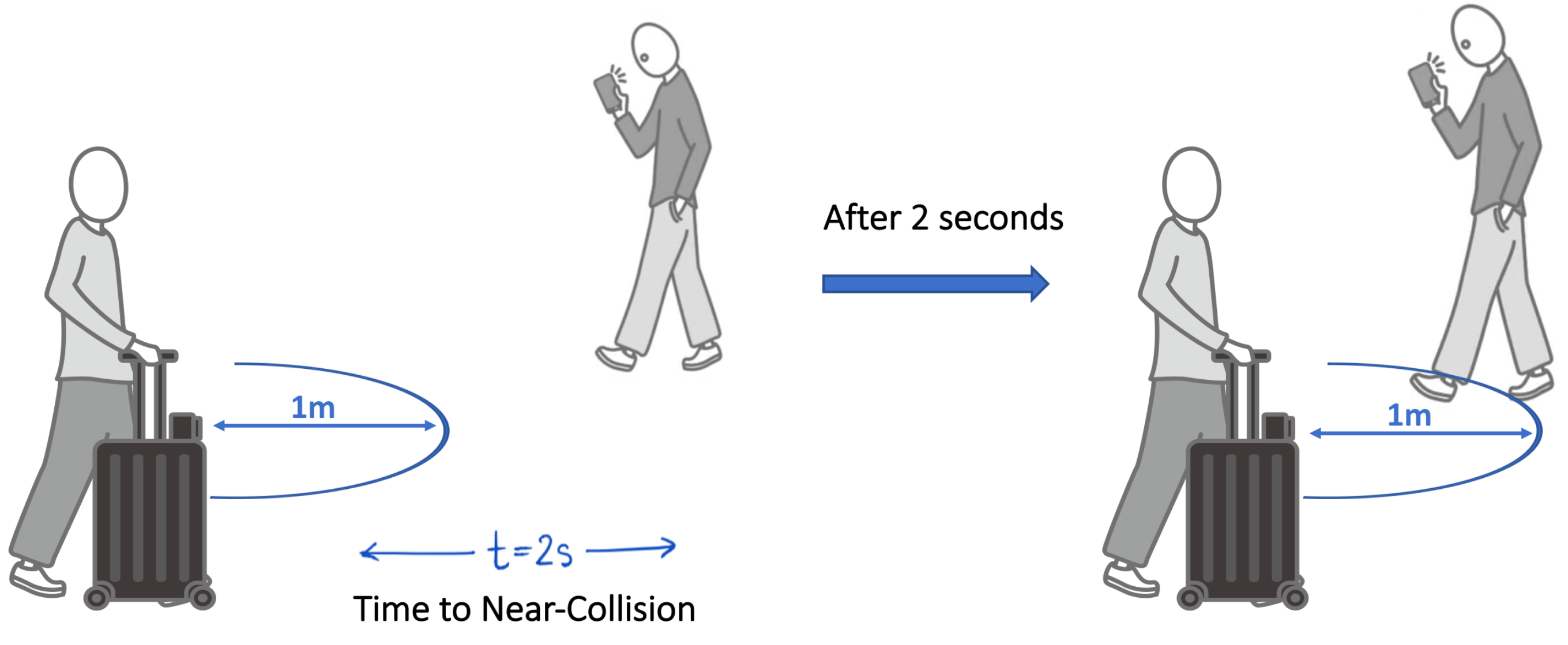}
      \vspace{-0.8cm}
      \caption{Forecasting time to near-collision between a suitcase-shaped robot being pushed by its user and nearby pedestrian.}
      \label{fig:illustration}
      \vspace{-0.35cm}
   \end{figure}

%%%%%%%%%% Second paragraph - Anticipating common arguments %%%%%%%%%%%%%%%%%%%
%A method frequently used for forecasting time to collision \cite{SIPP, BBeep} is to detect \cite{Ren2015, yolov3, Lee2016, Weng20182, Weng2019} the location of the surrounding pedestrians and track \cite{Bewley2016, Sharma2018, Weng2019_3dmot} their trajectories using a constant velocity model. 
% extrapolation is done using constant velocity model. I can add the detection citations later.
A method frequently used \cite{SIPP, BBeep} for forecasting time to collision is to track \cite{Andreas, Bewley2016, Sharma2018, Weng2019_3dmot} the 3D location of the surrounding pedestrians and extrapolate their trajectories using a constant velocity model. When it is possible to use high quality depth imaging devices, this type of physics-based modeling can be very accurate. However, physics-based approach can also be prone to failure in the presence of sensor noise and uncertainty in detection of nearby pedestrians. Small mistakes in the estimation of depth (common to low-cost depth sensors) or noise in 2D bounding box detection (common to image-based object detection algorithms) can be misinterpreted to be very large changes of velocity. Many physics-based approaches can be brittle in the presence of such noise. Accordingly, errors in either pedestrian detection, tracking or data association can result in very bad future trajectory estimates. Other more advanced physics-based models and decision-theoretic models also depend heavily on accurate state estimates of nearby people and can be significantly influenced by sensor and perception algorithm noise. We propose to address the issue of sensor noise and perception algorithm noise by directly estimating the time to collision from a sequence of images. Using a monocular camera is an alternate solution to using inexpensive RGB-D cameras like Kinect which are inapplicable for outdoor use.

To create a dataset for learning time to collision, we designed a training prototype that facilitates efficient annotation of large amounts of data. It is both unnatural and infeasible to record or insist that people actually collide with the mobile platform to collect large scale data. As an abstraction, we define the presence of a person within a 1 meter radius around the mobile platform as a near-collision. If a person is present within this radius, we mark it as a near-collision that should be forecasted using an earlier segment of video. The proposed approach is designed for a mobile robot that is being pushed by a person with visual impairment as shown in Fig. \ref{fig:illustration}. The goal of the system is to forecast the time to near-collision, few seconds before the near-collision event. While most of the existing datasets for human trajectory prediction are from a fixed overhead camera \cite{2009YoullNW}, \cite{UCY}, our dataset of \textbf{13,658} video segments targets the first-person view which is more intuitive for mobile robots. In the work on robust multi-person tracking from mobile platforms \cite{Andreas}, the dataset is egocentric at walking speed but with 4788 images it is insufficient to deploy the success of deep learning.

%%%%%%%%% Fifth paragraph - Technical part, how exactly we are learning %%%%%%%%%%%%%%%

We formulate the forecasting of time to near-collision as a regression task. 
%The prediction horizon is chosen to be 6 seconds for two reasons. First, at an average walking speed of {1.4 m/s}, predictions farther than 6 seconds are largely redundant in dynamic public places. Second, a planning budget of 6 seconds is enough to compute a collision avoidance maneuver. 
To learn the mapping from spatial-temporal motion of the nearby pedestrians to time to near-collision, we learn a deep network which, takes a sequence of consecutive frames as input and outputs the time to near-collision. To this end, we evaluate and compare two popular video network architectures in the literature: (1) The high performance of the image-based network architectures makes it appealing to reuse them with as minimal modification as possible. Thus, we extract the features independently from each frame using an image-based network architecture (\emph{e.g.}, VGG-16) and then aggregate the features across the temporal channels; (2) It is also natural to directly use a 3D ConvNet (\emph{e.g.}, I3D \cite{i3d}) to learn the seamless spatial-temporal features.
% To learn a task from a sequence of consecutive frames, we can either leverage the prior work in image domain or video. In image domain, we have 2D ConvNets changing from AlexNet \cite{alexnet} to ResNet \cite{resnet} that operate on a single image. One of the simple ideas is to extract the spatial features of each image in a sequence and then concatenate for temporal reasoning. 
% Another alternative is to operate directly on a video and extract seamless spatio-temporal features using inflated 3D ConvNet (I3D) \cite{i3d}. 
Moreover, it is a nontrivial task to decide how many past frames should form the input. Thus, we do extensive experimentation on different temporal windows as input using aforementioned video network architectures. Our results show that our proposed multi-stream CNN trained on the collected dataset is the best model for predicting time to near-collision.

%%%%%%%%%%%%%%%%%%%%%%%%%%%%%%%%%%%%%%%%%%%%%%%%%%%%%%%%%%%%%%%%%%%%%%%%%%%%%%%%%%%%%%%%

% The next section outlines the existing work in three closely related topics - deep learning for collision avoidance, human trajectory prediction and learning a task from videos. Section \ref{dataset} gives an overview of our proximity dataset. Section \ref{approach} describes the proposed network architecture to regress the time to proximity followed by Section \ref{experiments} reporting the performance of our model in comparison to existing approaches. 

In summary, the contributions of our work are as follows: (1) We contribute a large dataset of 13,658 egocentric video snippets of humans navigating in indoor hallways. In order to obtain ground truth annotations of human pose, the videos are provided with the corresponding 3D point cloud from LIDAR; (2) We explore the possibility of forecasting the time to near-collision directly from monocular images; (3) We provide an extensive analysis on how current state-of-the-art video architectures perform on the task of predicting time to near-collision on the proposed dataset and how their performance varies with different temporal windows as input.

% Summary of the contributions 

% The pedestrians are detected in an image using Faster-RCNN \cite{fasterRCNN} and their spatial location is computed by projecting the pixels within bounding box onto the corresponding 3D point cloud. Since the video is recorded at 10fps, the time of proximity is also extracted upto the resolution of 0.1 second.

%It should be noted that the LIDAR was only used to extract the ground truth and is not needed during inference.

%For the purpose of data augmentation, we used a stereo camera to double the amount the training data. An image of the setup used to collect data is shown in figure \ref{fig:setup}.  Using this setup, we collected a dataset of 14,000 image sequences each with having a time duration of 6.5 seconds - 0.5 seconds of history and 6 seconds for prediction.

%% file: sections/related_work.tex
\noindent
\textbf{Monocular-Based Collision Avoidance.} Existing monocular collision avoidance systems mostly focus on avoiding the immediate collision at the current time instant. Learning to fly by crashing \cite{gandhi} presented the idea of supervised learning to navigate an unmanned aerial vehicle (UAV) in indoor environments. The authors create a large dataset of UAV crashes and train an AlexNet \cite{alexnet} with single image as input to predict from one of these three actions - go straight, turn left or turn right. Similarly, DroNet \cite{DroNet} trains a ResNet-8 \cite{resnet} to safely navigate a UAV through the streets of the city.
% in contrast to traditional map-localize-plan approach. 
DroNet takes the images from an on-board monocular camera on UAV and outputs a steering angle along with the collision probability. 

While predicting an action to avoid the immediate collision is useful, it is more desirable to predict a possible collision in the short future.
To anticipate traffic accidents, Kataoka et al. \cite{near-miss} constructed a near-miss incident database of traffic scenes annotated with near-miss incident duration. By passing a video input to a quasi-recurrent neural network, their approach outputs a probability that an accident will occur in future. On an average, their approach is shown to anticipate a near-miss incident or accident 3.65 seconds in advance. While their study explored how early can a collision be anticipated, the exact time to collision was not predicted.
% Both of these papers focus on detecting immediate collision at current time instant and executing the required action - one formulates the output as a classification task on discrete actions whereas the latter regresses the continuous steering angle. 
Therefore, our work focuses on forecasting the exact time to a possible collision occurring within next 6 seconds, which will be helpful for collision avoidance in the context of dynamic path planning \cite{SIPP}, \cite{time-bounded}, \cite{vemula2016path}.   
% This prediction is useful in collision avoidance system designed for assistive robots supporting blind people \cite{BBeep} which uses preemptive sound notification to alert both the user and nearby pedestrians about the potential collision. 
% This will be helpful for planners \cite{time-bounded}, \cite{SIPP} that incorporate a time dimension only when a moving obstacle needs attention while navigating in dynamic environments.   

\vspace{2mm}\noindent\textbf{Predicting Time to Collision by Human Trajectory.}
Instead of predicting the time to collision directly from images, one can also predict the human trajectories \cite{socialGAN, socialLSTM, anirudh, 2009YoullNW, ziebart2009planning} as the first step, then the time to collision can be predicted based on the speed and heading directions of all the surrounding pedestrians.
% \textcolor{red}{need brief explanation for each paper here.}
\cite{2009YoullNW} introduces a dynamic model for human trajectory prediction in a crowd scene by modeling not only the history trajectory but also the surrounding environment. \cite{socialLSTM} proposes a LSTM model to learn general human movement pattern and thus can predict better future trajectory. As there are many plausible ways that humans can move, \cite{socialGAN} proposes to predict diverse future trajectories instead of a deterministic one. \cite{anirudh} proposes an attention model, which captures the relative importance of each surrounding pedestrian when navigating in the crowd, irrespective of their proximity.

However, in order to predict future trajectory reliably, these methods rely on accurate human trajectory history. This usually involves multi-people detection \cite{Ren2015, yolov3, Weng20182, Weng2019} and tracking, and thus has two major disadvantages: (1) Data association is very challenging in crowded scenarios. Small mistakes can be misinterpreted to be very large changes of velocity, resulting in very bad trajectory estimate; (2) Robust multi-person tracking from mobile platform is often time-consuming. For example, \cite{Andreas} takes 300ms to process one frame on a mobile GPU, making it impossible to achieve real-time collision forecasting. 

In contrast, our approach can predict the time to collision directly from a sequence of images, without requiring to track the surrounding pedestrians explicitly. We demonstrate that our data-driven approach can implicitly learn reliable human motion and also achieve real-time time to collision.
% The proposed multi-stream CNN overlay the spatial features extracted from one frame over the features of previous frame and thus implicitly learns to track people.

% You'll Never Walk Alone \cite{2009YoullNW} emphasized that the human trajectory prediction should take into account remaining scene objects but incorporated only the social interactions within people and their orientation towards the goal anticipated from the overhead camera. On the other hand, our approach takes into consideration the complete scene and thus it can exploit both the positions of scene objects such as walls and the motion of multiple pedestrians.
%Using multiple frames as input instead of a single image, our proposed method can learn the motion of persons relative to camera. 

\vspace{2mm}\noindent\textbf{Learning Spatio-Temporal Feature Representation.} Existing video architectures for spatio-temporal feature learning can be split into two major categories. To leverage the significant success from image-based backbone architectures (\emph{e.g.}, VGG and ResNet) pre-trained on large-scale image datasets such as ImageNet \cite{imagenet} and PASCAL VOC \cite{pascalVOC}, methods in the first category reuse the 2D ConvNet to extract features from a sequence of images with as minimal modification as possible. For example, \cite{BeyondSS} proposes to extract the image-based features independently from each frame using GoogLeNet and then apply a LSTM \cite{lstm} on the top for feature aggregation for video action recognition.

% explore different 3D ConvNet as it is natural to 

% The availability of very large image classification datasets such as ImageNet and PASCAL VOC \cite{pascalVOC} which led to significant improvements in deep architectures makes it appealing to use 2D ConvNets with minimal change required for video. 

% A theoretically sound approach to capture temporal dependencies is to add a recurrent layer such as LSTM on the top of spatial features extracted from 2D ConvNet \cite{BeyondSS}.

% However, recurrent neural networks are practically difficult to train and might not be needed for short snippets of video. 

The second category methods explore the use of 3D ConvNets for video tasks \cite{C3D, videoResNet, p3d, two-stream, Weng2019_lipreading} that directly operate 3D spatio-temporal kernels on video inputs. While it is natural to use 3D ConvNets for spatio-temporal feature learning, 3D ConvNets are unable to leverage the benefits of ImageNet pretraining easily and often have huge number of parameters which makes it most likely to overfit on small datasets. Recently, the two-stream inflated 3D ConvNet (I3D) \cite{i3d} is proposed to mitigate these disadvantages by inflating the ImageNet-pretrained 2D weights to 3D. Also, the proposed large-scale video dataset, Kinetics, has shown to be very successful for 3D kernel pre-training.
% performs better than the former approaches \cite{videoResNet}, \cite{p3d}, \cite{two-stream}. 
% I3D has lesser parameters than 3D ConvNet and also exploits ImageNet-pretrained 2D weights by inflating the Inception V1 \cite{inceptionv1} network. Two different versions of I3D which use a single stream - RGB-I3D and Flow-I3D with Kinetics pretraining, also have comparable accuracy on video classification. 

To validate how current state-of-the-art video-based architectures perform on the novel task of predicting time to near-collision on the proposed dataset, we evaluate methods from both categories in our experiments.

% Can RGB-I3D be used for collision forecasting from egocentric video? We will come back to this question in section \ref{experiments} of this paper.

%Video architectures such as ConvNet+LSTM and 3D ConvNet \cite{I3D} seem like a natural approach to learn the task of predicting time to collision from monocular video. Recent successes on two-stream networks showed the  

%% What is the idea behind I3D?

%% I transferred it to inroduction %%
%To learn a task from a sequence of consecutive frames, we can either leverage the prior work in image domain or video. In image domain, we have 2D ConvNets changing from AlexNet \cite{alexnet} to ResNet \cite{resnet} that operate on a single image. One of the simple ideas is to extract the spatial features of each image in a sequence and then concatenate for temporal reasoning. Another alternative is to operate directly on a video and extract seamless spatio-temporal features using 3D ConvNet or inflated 2D ConvNet I3D \cite{i3d}. 3D ConvNet has huge number of parameters and is most likely to overfit on our dataset of 13,685 examples. On the other hand, I3D is smaller with lesser parameters and also exploits ImageNet-pretrained 2D weights. I3D has been shown to perform the task of activity recognition from videos. Can the same network be used for collision forecasting from first person video? We will come back to this question in section \ref{experiments} of this paper.  

%% file: sections/dataset.tex
% \begin{figure}
%     \centering
%     \includegraphics[scale=0.40]{figs/setup.jpg}
%     \caption{Suitcase-shaped mobile system with stereo camera and LIDAR}
%     \label{fig:setup}
% \end{figure}

%%%%%%%%% Feb 18 %%%%%%%%%

\begin{table*}[ht] 
\centering
\caption{Public Video Datasets with Egocentric Viewpoint}\label{tab:data}
%\begin{tabular}{|c|c|c|c|} \hline
\vspace{-0.2cm}
\begin{tabular}{|P{4.1cm}|P{3cm}|P{2.2cm}|P{2cm}|P{4cm}|} \hline
Dataset & Number of near-collision video sequences & Structure of scenes & Setup for recording & Modalities Available \\ \hline
\textbf{Ours} (Near-collision)  & 13,685 & Indoor hallways & Suitcase & Stereo Depth + 3D Point Cloud \\ \hline 
UAV crashing \cite{gandhi} & 11,500 & Indoor hallways & UAV  & Monocular \\ \hline  
DroNet \cite{DroNet} & 137 & Inner-city & Bicycle & Monocular \\ \hline 
%Collision trajectory \cite{ziebart} & 166 & Kitchen Area & Laser range finders & Fixed \\ \hline  
Robust Multi-Person Tracking \cite{Andreas} & 350  & Busy inner-city & Chariot & Stereo Depth\\ \hline   
\end{tabular}
\vspace{-0.2cm}
\end{table*}

We sought to analyze a large-scale, real-world video dataset in order to understand challenges in prediction of near-collision events. However, based on our survey, existing datasets had a small number of interaction events as reported in Table \ref{tab:data} and lacked diversity in the capture settings. Therefore, in order to train robust CNN models that can generalize across scenes, ego-motion, and pedestrian dynamics, we collected an extensive dataset from a mobile perspective. Next, we describe our hardware setup and methodology for data collection. 
%We further provide the algorithm used for automatic ground truth annotation. 

\subsection{Hardware Setup}
\begin{figure}[t]
      \centering
      \includegraphics[height=3.8cm, width=\columnwidth]{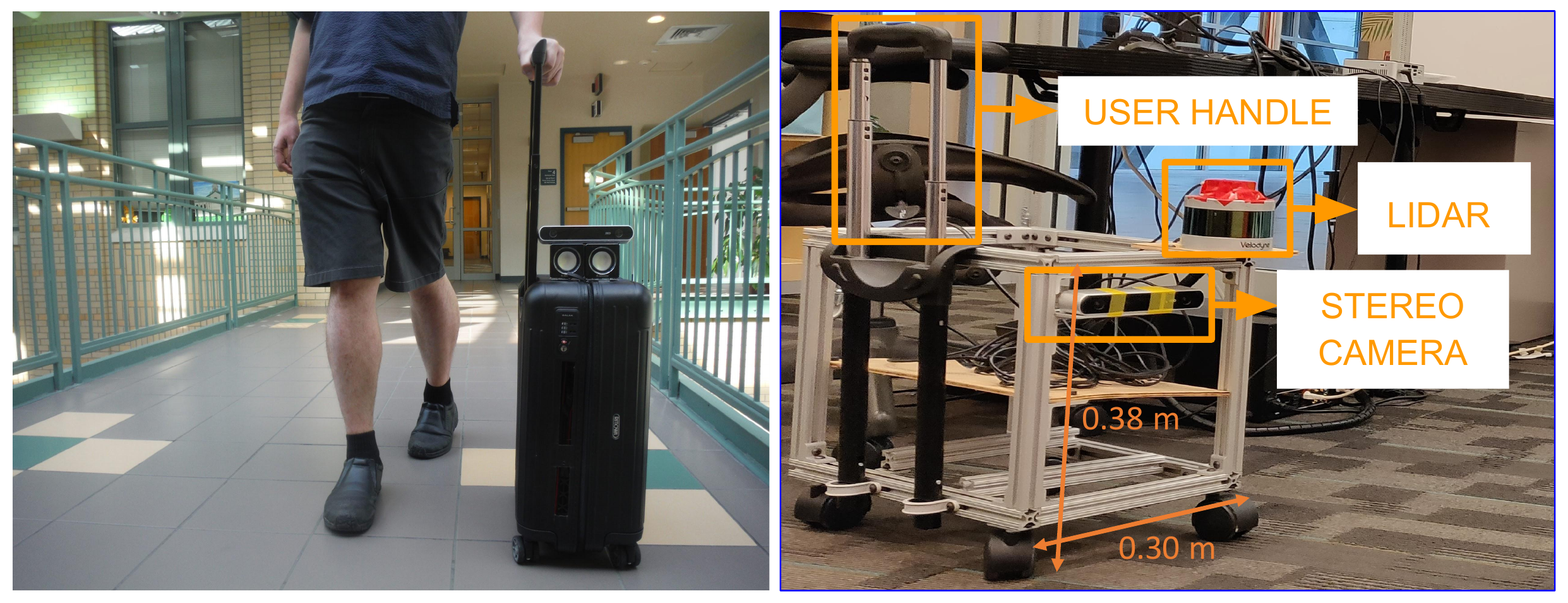}
      \vspace{-0.8cm}
      \caption{\textbf{Left}: We show an assistive suitcase with a camera sensor and speaker to guide people. \textbf{Right}: We show the corresponding suitcase-shaped training prototype mounted with stereo camera and LIDAR for data collection.}
      \label{fig:setup}
      \vspace{-0.2cm}
   \end{figure}

The mobile platform used for data collection is shown in Fig.~\ref{fig:setup}, and includes a stereo camera and LIDAR sensor. While during inference we only utilize a monocular video, the stereo camera serves two purposes. First, it provides a depth map to help in automatic ground truth annotation. Second, it doubles the amount of training data by providing both a left and right image perspective which can be used as separate training samples. However, during the process of automatic data annotation, it was observed that the depth maps from stereo camera are insufficient for extracting accurate distance measurements. In particular, when the pedestrian is close to camera the depth values are missing at corresponding pixels due to motion blur. To ensure stable and reliable depth annotations, we utilize a LIDAR sensor which is accurate to within a few centimeters. The images and corresponding 3D point clouds are recorded at the rate of 10Hz.     

% The mobile platform shown in \ref{fig:setup} is used to collect videos inside university buildings. The data is split into train and test as follows:
% \begin{itemize}
% \item Training images - 10685 (2106 positive and 8579 negative)  
% \item Test images - 3563 (687 positive and 2876 negative)
% \end{itemize}
%  The camera and LIDAR are calibrated using \cite{calibration}. We used Faster RCNN object detector to detect the pedestrian in RGB image and recorded the ground truth position in world from projected point cloud.

\subsection{Camera-LIDAR Extrinsic Calibration}
We use the camera and the LIDAR for automatic ground truth label generation. The two sensors can be initially calibrated with correspondences~\cite{Autoware,calibration}. An accurate calibration is key to obtaining the 3D position of surrounding pedestrians and annotating the large number of videos in our dataset. Let $R$ and $t$ denote the rotation matrix and the translation vector defining the rigid transformation between the LIDAR to the camera frame and $K$ the $3 \times 3$ intrinsic matrix of camera. Then, the LIDAR 3D coordinates $(x, y, z)$ can be related to a pixel in the image with coordinates $(U,V) = (\frac{u}{w}, \frac{v}{w})$ using following transformation:
%U = \frac{u}{w}; V = \frac{v}{w}
\begin{equation}\label{eq:Rt}
    \begin{bmatrix}u \\ v \\ w\end{bmatrix} = K[R \mid -R^{T}t] \begin{bmatrix} x \\ y \\ z \\ 1\end{bmatrix}
\end{equation}

Given this calibration, we can now project LIDAR points onto the image and obtain estimated depth values for the image-based pedestrian detection. 

%$$
%$$
%Here, . 
%% An example of projection of point cloud onto the image is shown in figure \ref{fig:projection}.
%% Show the pipeline - image showing projected point cloud, heatmap from GRADCAM 
%% Number of training and test images 

% \begin{figure}[h]
%     \centering
%     \includegraphics[scale=0.35]{figs/point_cloud_projected_over_image.eps}
%     \caption{Point cloud projected on image to extract 3D pose}
%     \label{fig:projection}
% \end{figure}

%For people detection, we employ a state-of-the-art person detection (Faster-R-CNN \cite{fasterRCNN}). Given a person detection, we can then compute its 3D location in the scene from the corresponding 3D point cloud.

%However, Autoware requires 30-40 poses of calibration target to output accurate transformation and thus we later shifted to \cite{calibration} which only requires 1-3 poses.  

\subsection{Data Collection and Annotation}
\begin{figure}
    \centering
    \includegraphics[height=5cm,width=\columnwidth]{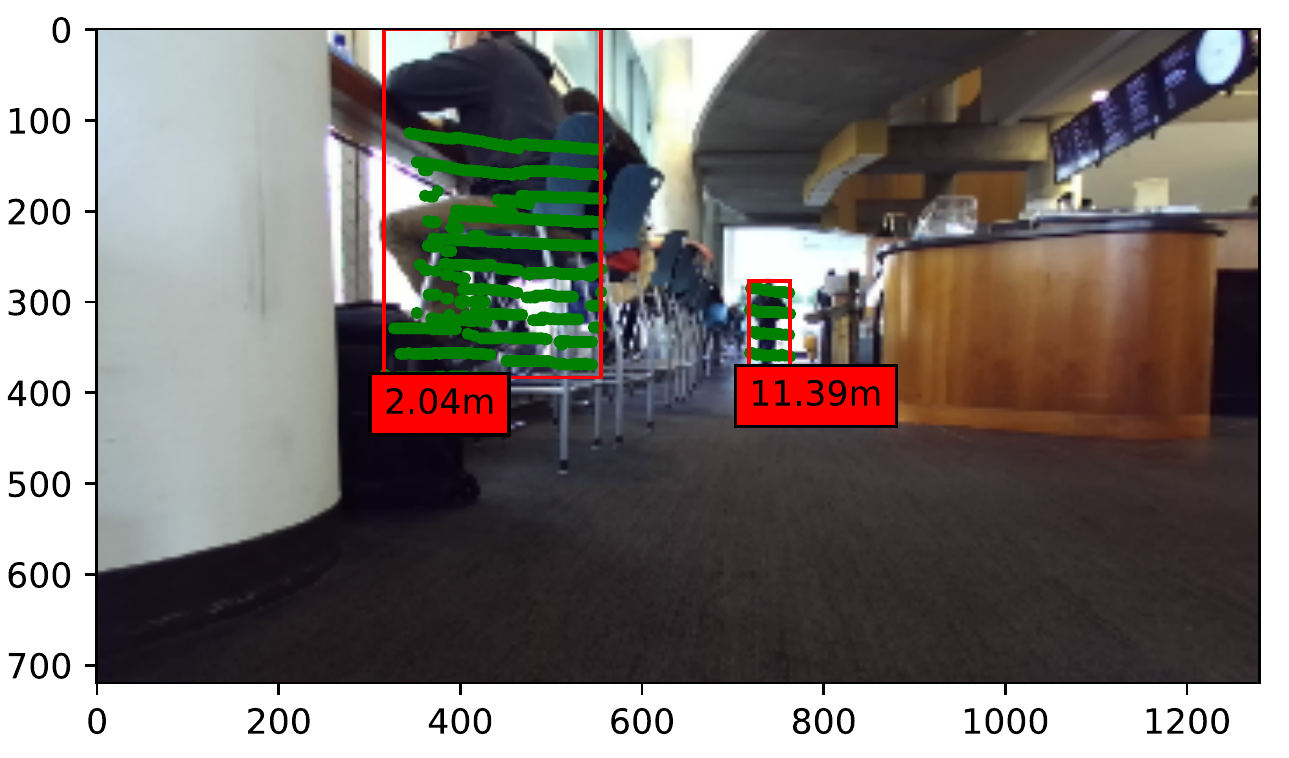}
    % \end{subfigure}
    \vspace{-0.8cm}
    \caption{Multi-modal ground truth generation. The two red bounding boxes indicate people detected by Faster R-CNN. The green points are projected to the image from the LIDAR, with the relative distance between LIDAR and person shown as well.}\label{fig:gt}
    \vspace{-0.2cm}
\end{figure}

The platform is pushed through three different university buildings with low-medium density crowd. Our recorded videos comprise of hallways of varying styles. 
%Each image is automatically annotated with a binary label using algorithm \ref{algo}. 
%An illustration of the resulting processing is shown in Fig. \ref{fig:gt}. A positive binary label indicates the presence of humans within the radius of 1 meter around the setup. As presented in line 3 of algorithm \ref{algo}, we detect the persons' bounding boxes by applying Faster RCNN \cite{fasterRCNN} on image and thus the persons behind the setup are not considered. 
We experimented with several techniques for obtaining pedestrian detections \cite{Ren2015, yolov3} in the scene from the image and LIDAR data. As 2D person detection is a well-studied problem, we found an image-based state-of-the-art person detection (Faster-R-CNN \cite{Ren2015}) to perform well in most cases, and manually inspect and complete any missing detections or false positives. To obtain the 3D position of each detected bounding box, we compute a median distance of its pixels using the 3D point cloud. An illustration of the resulting processing is shown in Fig. \ref{fig:gt}. Each image is annotated with a binary label where a positive label indicates the presence of at least one person within a meter distance from setup. We understand that some people in immediate vicinity moving in the same direction as camera might not be important for collision, but due to close proximity of 1 meter should still be recognized and planned over carefully.
%However, the number of such instances is insignificant in our dataset. \\

%Some of the halls also have students sitting on study tables who are also referred to as pedestrians in this paper.

% \begin{algorithm}
% \caption{Ground Truth Labeling}
% \label{algo}
% \begin{algorithmic}[1]
% \STATE \text{\textbf{Input:} A tuple of ($image_t, LIDAR_t$)}
% \STATE \text{\textbf{Init:} $label_{t} = 0$}
% \STATE \text{Bounding boxes $\leftarrow$ FasterRCNN($image_t$)}
% \FOR {\text{box in Bounding Boxes}}
% \STATE \text{distance inside box = $\emptyset$}
% \FOR {\text{point $[x, y, z]$ in $LIDAR_{t}$}}
% \STATE $\begin{bmatrix}u \\ v \\ w\end{bmatrix} = K[R \mid -R^{T}t] \begin{bmatrix} x \\ y \\ z \\ 1\end{bmatrix}$ \COMMENT{from Eq. \ref{eq:Rt}}
% \IF {$(\frac{u}{w}, \frac{v}{w})$ lies inside box}
% \STATE \text{add $\sqrt{x^{2} + y^{2}}$ to distance inside box}
% \ENDIF
% \ENDFOR
% \STATE \text{pedestrian distance = Median(distance inside box)}
% \IF {pedestrian distance $<$ 1 meter}
% \STATE \text{$label_{t} = 1$}
% \ENDIF
% \ENDFOR
% \STATE \text{return $label_{t}$}
% \end{algorithmic}
% \end{algorithm}

Now we want to estimate the time to near-collision, in terms of milliseconds, based on a short temporal history of few RGB frames. Let us consider a tuple of $N$ consecutive frames 
%$(image_{1}, image_{2}, \hdots, image_{N})$ 
$(I_{1}, I_{2}, \hdots, I_{N})$ and using this sequence as history we want to estimate if there is a proximity over the next 6 seconds. 
%Predictions farther than 6 seconds will largely be redundant while navigating in dynamic public places. 
Since the framerate is $10$ fps, we look at the next 60 binary labels in future annotated 
%using algorithm \ref{algo}
as $\{label_{n+1}, label_{n+2}, \hdots, label_{n+60}\}$. If we denote the index of first positive label in this sequence of labels as $T$ then our ground truth time to near-collision is $t = \frac{T}{10}$ seconds. 

\subsection{Comparison with Existing Datasets}
In Table \ref{tab:data}, we compare our proposed dataset with existing datasets recorded from egocentric viewpoint in terms of (1) number of near-collision video sequences, (2) structure of scenes, and (3) setup used for recording. UAV crashing \cite{gandhi} dataset is created by crashing the drone 11,500 times into random objects. DroNet \cite{DroNet} has over 137 sequences of starting far way from an obstacle and stopping when the camera is very close to it. The two main reasons for collecting proposed dataset over existing datasets of UAV crashing and DroNet are: (1) applicability to assistive suitcase system \cite{BBeep}, and (2) focus on pedestrian motion. While the dataset provided by Ess et al \cite{Andreas} suited to our application, we find only $350$ near-collision instances making it infeasible to deploy CNNs.

%% file: sections/approach.tex
  \begin{figure*}[ht]
      \centering
      \includegraphics[height=7.0cm, width=12cm]{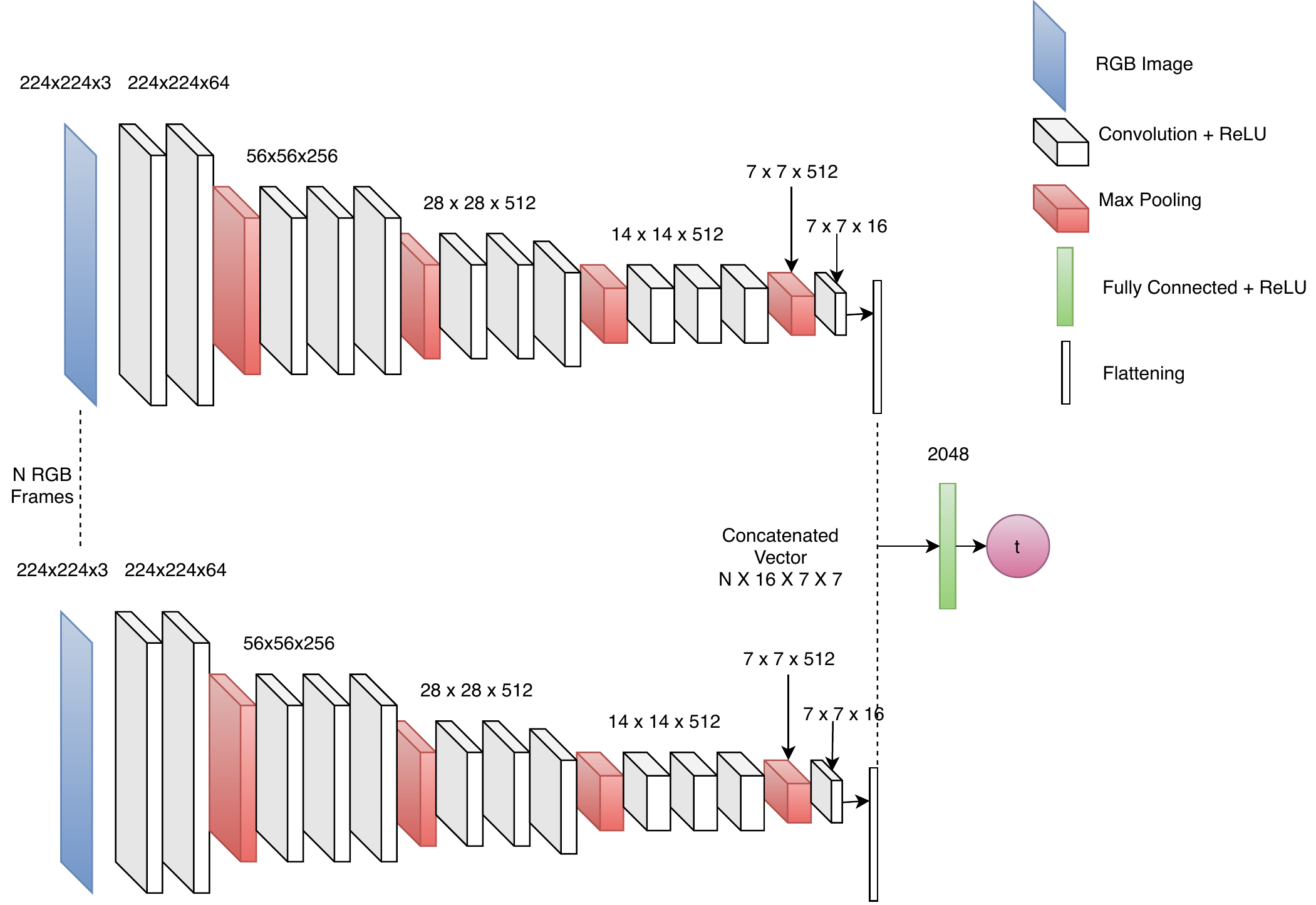}
      \vspace{-0.2cm}
      \caption{Our model with VGG-16 as backbone where the final output is time to collision denoted as $t$ }
      \label{fig:model}
  \end{figure*}

Our goal is to predict the time at which at least one person is going to come within a meter distance of the mobile setup using only a monocular image sequence of $N$ frames. The video is recorded at $10$ fps and thus a sequence of $N$ frames, including the current frame and past $(N-1)$ frames, correspond to the history of $(N-1)/10$ seconds. We first provide a formal definition of the task and then the details of network architecture for reproducibility.   

\subsection{Problem Formulation - Classification or Regression?}
Learning time to near-collision can be formulated as a multi-class classification into one of the 60 classes where $i^{th}$ class corresponds to time range between $((i-1)/10, i/10]$ seconds. The disadvantage of training it as a classification task is that all the mispredictions are penalized equally. For example, let us consider two different mispredictions given the same ground truth of 0.5 seconds - one where the network categorized it into the class $(0.6, 0.7]$ and other when the network predicted $(5.5, 5.6]$. The multi-class cross-entropy loss on both of these will be equal while we want the latter to be penalized much more than the former. One of the solutions is to design a differentiable loss function which keeps this preference in mind. Another solution is to formulate it as a regression problem and use the mean-squared error as the loss function. In this paper, we formulated it as a regression problem as follows:
$$
%t = f(image_{1}, image_{2}, \hdots, image_{N}) \text{ where } t \in [0, 6]
t = f(I_{1}, I_{2}, \hdots, I_{N}) \text{ where } t \in [0,6]
$$

\subsection{Network Architecture}
VGG-16 \cite{vgg} is a 16-layer convolutional neural network which won the localization task in ImageNet Challenge 2014. It used parameter efficient $3 \times 3$ convolutional kernels pushing the depth to 16 weight layers. It was shown that its representations generalize well to other datasets achieving state-of-the-art results. We propose a multi-stream VGG architecture as shown in Fig. \ref{fig:model} where each stream takes a $224 \times 224$ RGB frame as input to extract spatial features . These spatial features are then concatenated across all frames preserving the temporal order and then fed into a fully-connected layer to output time to collision. \\

% The inputs to the network are short N-frame clips corresponding to a temporal footprint of $\frac{N-1}{10}$ seconds. The concatenated features are fed into a 2 layer perceptron to give a single real-valued output. The network is trained using mean squared error as loss on the output. \\

\textbf{Feature Extraction from VGG-16} We extracted the features of dimensions  7$\times$7$\times$512 from the last max pool layer of VGG-16. These features pass through an additional convolution layer to reduce the feature size to 7$\times$7$\times$16 and then flattened. These flattened features for each frame are concatenated into a vector and fed into the successive fully-connected layer of size 2048 which finally leads to a single neuron denoted as $t$ in Fig. \ref{fig:model}.  
%The network is trained using mean squared error as loss on the output.    \\
%% Which layer of VGG-16; I also added a conv layer from 512 channels to 16. 

In this network, the convolutional operators used spatial 2D kernels. A major question in current video architectures is whether these 2D kernels should be replaced by 3D spatio-temporal kernels \cite{i3d}. To address this question, we also experimented with 3D spatio-temporal kernels and report the results in following section. \\

\textbf{Training N-stream VGG} We initialized the VGG-16 network using ImageNet-pretrained weights. As the ImageNet dataset does not have a person class, we fine-tuned the network weights on PASCAL VOC \cite{pascalVOC} dataset. Using these weights as initialization, we train a multi-stream architecture with shared weights. The network is trained using the following loss function.
$$
L_{MSE} = \frac{1}{2}||t_{true} - f(I_1, I_2, \hdots, I_N)||^{2}
$$
Here, $L$ is the mean squared loss between the predicted time, \emph{i.e.}, $f(I_1, I_2, \hdots, I_N)$ and ground truth time denoted as $t_{true}$. \\
The loss is optimized using mini-batch gradient descent of batch size 24 with the learning rate of $0.001$. The training data is further doubled by applying horizontal flip transformation.
%% Leaning rate, gradient descent, batch size

%% file: sections/experimental_evaluation.tex
We now describe our evaluation procedure to decide the optimum temporal window as input on two different video network architectures. We further compare the performance with strong collision prediction baselines.

\subsection{Different temporal windows as input}
A single image can capture spatial information but no motion characteristics. Thus, we propose to use a sequence of image frames as history.  By feeding $N$ image frames, we consider a history of $(N-1)/10$ seconds. The temporal window of input frames was gradually increased from 2 frames (0.1 sec) to 9 frames (0.8 sec). To quantify the performance, we measure the mean absolute error (MAE) for the predictions on the test set and the standard deviation in error. From Table \ref{tab:hist}, it is empirically concluded to use a temporal window of 0.5 seconds, i.e, 6 frames in multi-stream VGG for most accurate predictions. 

% \begin{table}[h]
% \caption{Performance of N-stream VGG on proximity dataset vs number of input frames N}\label{tab:hist}
% \begin{tabular}{|c|c|c|c|} \hline
% Number of frames & Multi-stream VGG & I3D \\ \hline 
% 1 & 0.879  $\pm$  0.762 & 0.961 $\pm$ 0.707  \\ \hline
% 2 & 0.739 $\pm$  0.696 & 0.879 $\pm$ 0.665 \\ \hline 
% 3 & 0.701  $\pm$  0.622 & 0.914 $\pm$ 0.659 \\ \hline 
% 4  & 0.766 $\pm$  0.739 & 0.811 $\pm$ 0.642  \\ \hline
% 5 & 0.849 $\pm$  0.734 & 0.845 $\pm$ 0.658 \\ \hline 
% 6  & \textbf{0.753} $\pm$  \textbf{0.687} & 0.816 $\pm$ 0.663   \\ \hline
% 7 & 0.727 $\pm$  0.670 & 0.848 $\pm$ 0.733 \\ \hline
% 8 & 0.997 $\pm$ 0.875 & 0.811 $\pm$ 0.647 \\ \hline 
% 9 & 0.817 $\pm$  0.738 & 0.855 $\pm$ 0.670 \\ \hline
% \end{tabular}
% \end{table}

\begin{table}[ht]
\caption{Distribution of absolute error (mean $\pm$ std) on near-collision dataset using different number of frames}\label{tab:hist}
\begin{tabular}{|P{2cm}|P{2.5cm}|P{2.5cm}|} \hline
Number of frames & Multi-stream VGG  & I3D  \\ \hline 
1 & 0.879 $\pm$ 0.762s & 0.961 $\pm$ 0.707s  \\ \hline
2 & 0.828 $\pm$  0.739s & 0.879 $\pm$ 0.665s \\ \hline 
3 & 0.826  $\pm$  0.647s & 0.914 $\pm$ 0.659s \\ \hline 
4  & 0.866 $\pm$  0.696s & \textbf{0.811 $\pm$ 0.642s}  \\ \hline
5 & 0.849 $\pm$ 0.734 & 0.845 $\pm$ 0.658s \\ \hline 
6  & \textbf{0.753 $\pm$ 0.687s} & 0.816 $\pm$ 0.663s   \\ \hline
7 & 0.757 $\pm$  0.722s & 0.848 $\pm$ 0.733s \\ \hline
8 & 0.913 $\pm$ 0.732s & 0.811 $\pm$ 0.647s \\ \hline 
9 & 0.817 $\pm$  0.738s & 0.855 $\pm$ 0.670s \\ \hline
\end{tabular}
\vspace{-0.3cm}
\end{table}

\subsection{Experimental comparison of architectures}
We show a comparison of the performance of multi-stream VGG model and baselines including state-of-the-art methods in Table \ref{tab:baselines}. 

\begin{table}[ht]
\caption {Distribution of absolute error (mean $\pm$ std) on regression task compared with different baselines} \label{tab:baselines} 
\vspace{-0.2cm}
\begin{tabular}{|P{4cm}|P{1.5cm}|P{1.5cm}|} \hline
Method  &  Mean (in s) & Std (in s)\\ \hline
Constant baseline ($\mathbb{E}[y_{true}]$) & 1.382 & 0.839\\ \hline 
Tracking + Linear Model \cite{BBeep} &  1.055  & 0.962 \\ \hline 
DroNet \cite{DroNet} & 1.099 &  0.842  \\ \hline 
Gandhi et al \cite{gandhi} & 0.884 & 0.818 \\ \hline
Single Image VGG-16 & 0.879  & 0.762 \\ \hline
I3D (4 frames) \cite{i3d} & 0.811  & 0.642  \\ \hline
\textbf{Multi-stream VGG (6 frames)} & \textbf{0.753}  & \textbf{0.687}  \\ \hline 
\end{tabular}
\vspace{-0.2cm}
\end{table}

  \begin{figure}[ht]
      \centering
      \includegraphics[width=\columnwidth]{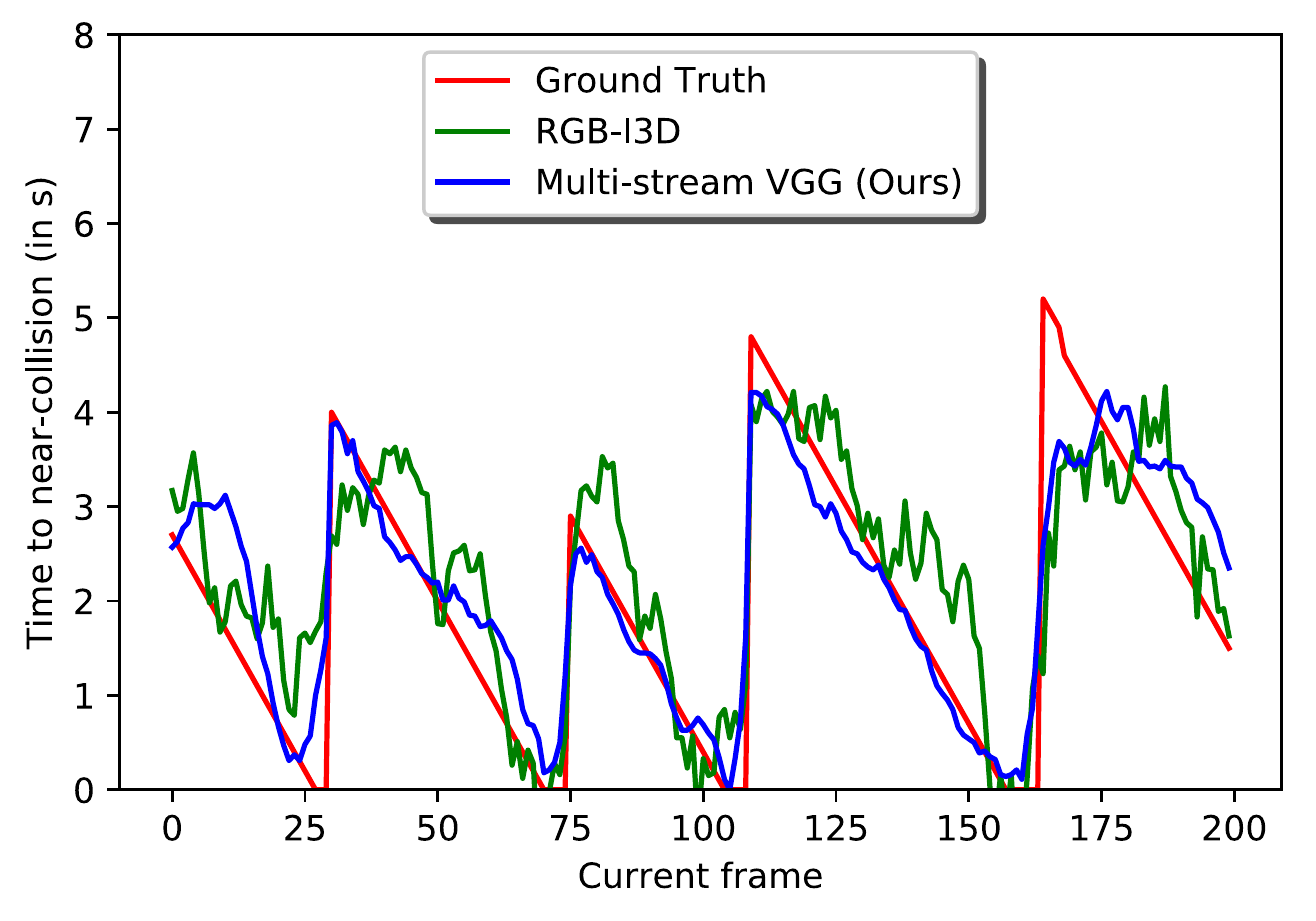}
      \vspace{-0.85cm}
      \caption{Example prediction results over consecutive frames. The discontinuities in the ground truth occur when the nearest pedestrian exits the view.
      As shown, the multi-stream VGG approach better adheres to the ground truth compared to I3D both when the pedestrian is far and near the camera, while providing smoother and more temporally-consistent predictions. }
      \label{fig:plot}
      \vspace{-0.3cm}
  \end{figure}
  
    \begin{figure*}[ht]
      \centering
      \includegraphics[height=7.5cm, width=\textwidth]{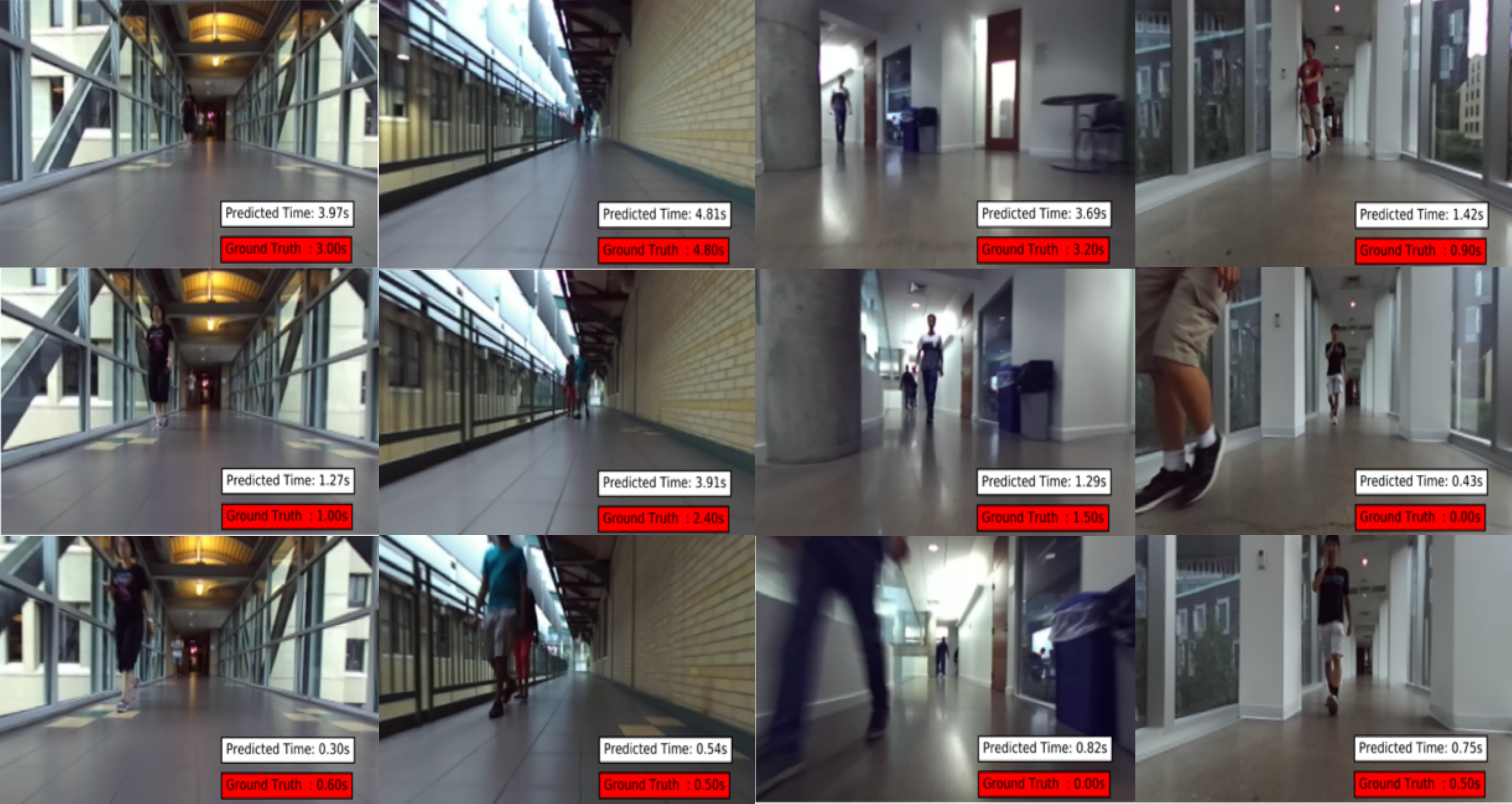}
      \vspace{-0.7cm}
      \caption{Predictions on four different test videos}
      \vspace{-0.2cm}
      \label{fig:testVideos}
  \end{figure*}

%     \begin{figure}[ht]
%       \centering
%       \includegraphics[height=7cm, width=\columnwidth]{figs/qualitative_results.pdf}
%       \caption{Qualitative results on theee different videos}
%       \label{fig:plot}
%   \end{figure}
%\begin{enumerate}
    % \textit{Constant Baseline:} 
    \subsubsection{Constant Baseline}
    On the training data of 12,620 samples, we compute the mean time to near-collision denoted by $\mathbb{E}[y_{true}]$ as a weak baseline. For each test input, we predict $\mathbb{E}[y_{true}]$ which was found to be $2.23$ seconds. 
    
    \subsubsection{Tracking followed by constant velocity model}
    %\item \textit{Tracking followed by constant velocity model:} 
    In dynamic environments, pedestrians are often tracked using a stereo camera or LIDAR. By saving few previous locations (0.5-2 seconds), a linear regression fit is used to predict the velocity vector of person \cite{BBeep, SIPP}. This velocity vector is then linearly extrapolated to predict where the pedestrian will be over the next 6 seconds and the corresponding accuracy is reported in Table \ref{tab:baselines}. A major disadvantage in this method is the need for image-based tracking which is less reliable at low framerate of $10$ fps.     
    
    \subsubsection{Deep learning for collision avoidance}
    %\item \textit{Deep learning for collision avoidance:} 
    Collision avoidance using deep learning has been previously proposed in \cite{gandhi} and \cite{DroNet}. Gandhi et al \cite{gandhi} created a UAV crash dataset to train AlexNet for classifying the current image into one of these three categories of drone policy: go straight, turn left or turn right. For learning time to near-collision using their approach, we take a single image as input and use AlexNet architecture with ImageNet-pretrained weights as initialization. The only difference lies in the output layer which is a single neuron regression instead of a three-neuron classifier. Our multi-stream VGG outperformed current-frame AlexNet as reported in Table \ref{tab:baselines}.
    
ResNet-8 architecture used in DroNet \cite{DroNet} takes in the single image and after the last ReLU layer splits into two fully-connected streams outputting steering angle and collision probability respectively. To experiment with their learning approach, we used ResNet-8 architecture with only one output, i.e., time to near-collision. The performance is close to the constant velocity prediction model as reported in Table \ref{tab:baselines} and thus it can be seen that it is unable to leverage real-world data. One of the reasons for its low performance could be the unavailability of ImageNet-pretrained weights for ResNet-8 architecture and thus it has to be trained from scratch on our dataset which is much smaller than the Udacity's car-driving dataset of 70,000 images used for training steering angle stream in DroNet.   
    
    %% should elaborate in related work 
    %% Here only the network architecture and why is it chosen as a baseline 
    \subsubsection{I3D for action classification in videos}
    %\item \textit{I3D for action classification in videos:} 
    Two-Stream Inflated 3D ConvNet (I3D) \cite{i3d} is a strong baseline to learn a task from videos. All the $ N \times N$ filters and pooling kernels in ImageNet-pretrained Inception-V1 network \cite{inceptionv1} are inflated with an additional temporal dimension to become $N \times N \times N$. I3D has two streams - one trained on RGB inputs and other on optical flow inputs. To avoid adding the latency of optical flow computation for real-time collision forecasting, we only used the RGB stream of I3D. We fine-tuned the I3D architecture which was pre-trained on Kinetics Human Action Video dataset \cite{i3d} on our near-collision dataset by sending $N$ RGB frames as input where $N = \{1,2, \hdots, 8,9\}$ as reported in Table \ref{tab:hist}. The outermost layer is modified from 400-neuron classifier to 1-neuron regressor. Since our $N$-frame input is smaller than the original implementation on 64-frame input, we decreased the temporal stride of last max-pool layer from 2 to 1. While 6 input frames were found to be the best for proposed multi-stream VGG network, we experimented again with the optimal history on I3D. The performance of I3D with varying number of input frames is reported in Table \ref{tab:hist}.  For $N = 4, 6, 8$, I3D is found to give the best results among 1-9 frames though our multi-stream VGG prediction for $N = 6$ outperformed the I3D prediction in best case. 
%\end{enumerate}
\subsection{Qualitative Evaluation}
% After seeing quantitative results as reported in Table \ref{tab:baselines}, the mean absolute error between proposed approach and the closest baseline I3D differs by a small number of 0.05 seconds. Does it mean that the proposed method offers only a slight improvement over existing approaches which might be negligible during real-time execution? To answer this we plotted the predictions made by proposed network and most competitive baseline I3D to compare with the ground truth. 
From the plots shown in Fig. \ref{fig:plot} we can observe that the predictions given by multi-stream VGG on 6 frames give smoother output as compared to undesired fluctuations in I3D output. We also qualitatively show in Fig. \ref{fig:testVideos} the comparison of time to near-collision predicted by our method vs the ground truth. 
%This motivates the introduction of another evaluation metric - smoothness factor. 

\subsection{Forecast Horizon}
In Table \ref{tab:error_vs_time_interval}, we report the error at different time-to-collision intervals. It is easiest to forecast the collision a second away and hence the error is least. In general, as forecast horizon increases it keeps on getting more difficult to forecast the exact time-to-collision.
\begin{table}[ht]
\caption{How error in prediction varies with time-to-collision?}\label{tab:error_vs_time_interval}
\vspace{-0.2cm}
\begin{tabular}{|P{2.4cm}|P{2.4cm}|P{2.4cm}|} \hline
Ground Truth Time-to-Collision Interval (in s) & Number of Test Samples & Mean Absolute Error in Predictions (in s)  \\ \hline
0-1 & 348 & 0.6027\\ \hline 
1-2 & 211 & 0.6446\\ \hline 
2-3 & 188 & 0.7547\\ \hline 
3-4 & 147 & 0.7189\\ \hline 
4-5 & 86 & 0.9502\\ \hline 
5-6 & 58 &  1.8369 \\ \hline 
\end{tabular}
\vspace{-0.3cm}
\end{table}